\documentclass[journal]{IEEEtran}
\usepackage{cite}
\usepackage[square,sort,comma,numbers]{natbib}

\ifCLASSINFOpdf
\usepackage[pdftex]{graphicx}
\else
\fi
\usepackage{amsmath}
\usepackage{amsfonts}
\usepackage{xcolor}
\usepackage{bbding}
\usepackage{multirow}
\usepackage{booktabs}
\usepackage{array}

\usepackage[caption=false,font=normalsize,labelfont=sf,textfont=sf]{subfig}
\usepackage{url}

\begin{document}
%
\title{Changes-Aware Transformer: Learning Generalized Changes Representation}
%
%
%

\author{Dan~Wang,~\IEEEmembership{Student~Member,~IEEE,}
	Licheng~Jiao,~\IEEEmembership{Fellow,~IEEE,}
	Jie~Chen,~
	Shuyuan~Yang,~\IEEEmembership{Senior~Member,~IEEE,}
	and~Fang~Liu,~\IEEEmembership{Senior~Member,~IEEE}
	\thanks{This work has been submitted to the IEEE for possible publication. Copyright may be transferred without notice, after which this version may no longer be accessible.}}

%
%

\markboth{Journal of \LaTeX\ Class Files,~Vol.~14, No.~8, August~2015}%
{Shell \MakeLowercase{\textit{et al.}}: Bare Demo of IEEEtran.cls for IEEE Journals}
%



\maketitle

\begin{abstract}
	Difference features obtained by comparing the images of two periods play an indispensable role in the change detection (CD) task. However, a pair of bi-temporal images can exhibit diverse changes, which may cause various difference features. Identifying changed pixels with differ difference features to be the same category is thus a challenge for CD. Most nowadays' methods acquire distinctive difference features in implicit ways like enhancing image representation or supervision information. Nevertheless, informative image features only guarantee object semantics are modeled and can not guarantee that changed pixels have similar semantics in the difference feature space and are distinct from those unchanged ones. In this work, the generalized representation of various changes is learned straightforwardly in the difference feature space, and a novel Changes-Aware Transformer (CAT) for refining difference features is proposed. This generalized representation can perceive which pixels are changed and which are unchanged and further guide the update of pixels' difference features. CAT effectively accomplishes this refinement process through the stacked cosine cross-attention layer and self-attention layer. After refinement, the changed pixels in the difference feature space are closer to each other, which facilitates change detection. In addition, CAT is compatible with various backbone networks and existing CD methods. Experiments on remote sensing CD data set and street scene CD data set show that our method achieves state-of-the-art performance and has excellent generalization.
\end{abstract}

\begin{IEEEkeywords}
	Change detection, Transformer, difference feature learning, cross-attention, self-attention.
\end{IEEEkeywords}

%
\IEEEpeerreviewmaketitle

\section{Introduction}
\begin{figure}[!t]
	\centering
	\includegraphics[width=2.5in]{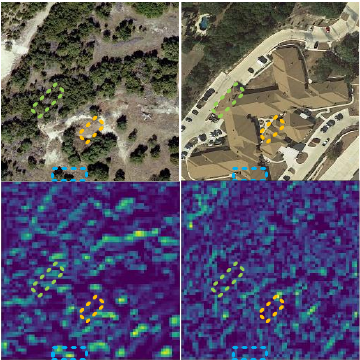}
	\caption{Various changes in bi-temporal figures.}
	\label{fig_diff_diff}
\end{figure}

%
%
%
%
\IEEEPARstart{C}{hange} detection aims to identify interested changes in two images of different periods of the same area (bi-temporal images) and give the pixel-level dense change map. Comparing the differences between the bi-temporal images is necessary to achieve this goal. Generally, change detection methods model the bi-temporal image features and then extract difference features from them. The final result is predicted based on the difference features. However, bi-temporal images often contain more than one kind of change, which may lead changed pixels to have very different differences features and cause troubles. Take Fig. \ref{fig_diff_diff} as an example. There are at least two kinds of changes: bare soil to buildings (the yellow box) and plants to buildings (the green box). The area in the blue box has both "bare soil-buildings" change and "plants-buildings" change. The figures in the second row show two difference feature maps obtained by the subtraction of the corresponding bi-temporal image features. The changed pixels of "bare soil-buildings" and those of "plants-buildings" are differ. Obtaining distinctive difference features of changed pixels that are as similar as possible from each other and as dissimilar as possible from those unchanged ones is the core challenge of CD.

To address this challenge and improve performance, many prior methods focus on modeling semantic-rich bi-temporal image representations \cite{rahman2018siamese, chen2020spatial, chen2020dasnet, peng2020optical, chen2021remote, zhou2022spatial}, improving supervision information \cite{zhang2020deeply, zhang2021escnet}, etc. Although informative bi-temporal image features are helpful for revealing changes, these implicit ways are hard to guarantee that different kinds of changed pixels have common difference features and will be classified into the same category. It raises the question: how to effectively make various changed pixels have common difference information to facilitate change detection?

In this article, we propose directly learning the generalized changes (GC) representation in the difference feature space and leveraging it to enhance pixels' difference features. We develop a novel Changes-Aware Transformer (CAT) based on the window-based attention Transformer, NesT \cite{zhang2022nested} to achieve this. The key idea is to embed the same change information into the changed pixels in the difference feature space to make them more similar, as well as more distinct from unchanged ones. Specifically, we first extract multi-scale bi-temporal image features to produce the initial difference feature maps (IDFs) and feed them to the CAT. Then, CAT learns the GC representation in the difference feature space by deep supervision. The GC representation collecting the knowledge of various change information can be seen as a known changed pixel. To embed the GC representation into changed pixels, we design a cosine cross-attention layer, and the GC representation provides explicit awareness of other changed pixels by calculating cosine similarity with every pixel. Intuitively, the greater the absolute value of the cosine similarity between a pixel and the GC representation, the more likely it is to be a changed pixel. Then, The GC representation updates pixels' difference features according to the cosine cross-attention weights. In this way, changed pixels' difference features are enhanced to varying degrees by the same generalized changes knowledge, and they could become more similar. A self-attention layer following the cross-attention layer makes pixels interact with and gather information from other pixels. CAT adopts the window-based attention architecture to save computing resources, and we perform all self-attention computations in the local window to make CAT run more efficiently.

The proposed CAT is a general method which can be integrated to various backbone networks or existing methods. Additionally, we develop a local window attention Transformer-based backbone and a dense upsampling module for CD. In general, our contributions are three-fold:
\begin{enumerate}
	\item We propose directly learning the generalized changes representation in the difference feature space for CD. The GC representation collects various change information of a pair of bi-temporal images and can explicitly perceive changed pixels and make them more distinctive. It provides a direct and effective way to enhance difference features.
	
	\item We develop a novel Changes-Aware Transformer (CAT) working in the difference feature space. The cosine cross-attention layer of CAT perceives and updates changed pixels using the GC representation in an efficient way. Through the updating, changed pixels can share common change information and become more similar to each other, as well as more distinct from unchanged pixels.
	
	\item CAT can be integrateted with Transformers, CNNs backbone, and existing methods. Experiment results confirm the models equipped with CAT achieve state-of-the-art (SOTA) performance on remote sensing CD data set (the LEVIR, DSIFN, and CDD data sets) and street scene CD data set (the VL-CMU-CD data set).
\end{enumerate}

\section{Related Work}
\subsection{Change Detection}
To obtain distinctive difference features, previous works mainly focused their efforts on three aspects:

\subsubsection{Improve the backbone network and enhance image representation}
\citet{zhan2017change} learned a deep Siamese convolution network \cite{chopra2005learning} to extract features from image pairs. \citet{rahman2018siamese} proposed the use of multi-scale features. \citet{chen2020spatial}, \citet{chen2020dasnet}, \citet{peng2020optical}, \citet{jiang2020pga}, \cite{zhou2022spatial} and \citet{chen2021remote} utilized the attention mechanism \cite{bahdanau2015neural, hu2018squeeze} to learn spatial, temporal, or channel context to enhance the image features. \citet{lei2021difference} employed the difference feature to learn channel attention to rescale bi-temporal image features and then influence final difference feature maps acquisition. \citet{bandara2022transformer} and \citet{zhang2022swinsunet} built the Siamese framework by means of Transformer to take advantage of the potential of Transformers in CV.

\subsubsection{Study the way of image feature fusion}
\citet{daudt2018fully} discussed two methods of acquiring difference features, absolute difference and concatenation of image features, based on a fully convolutional Siamese network. \citet{zhang2020deeply} combined spatial and channel attention to fuse image features to identify changes. \citet{lei2021difference} transformed concatenated image features into multi-scale features and then computed their global attention to remove redundant information in the concatenated features.

\subsubsection{Multi-task and Multi-supervisory information}
\citet{zhang2021escnet} introduced the superpixel segmentation algorithm and deep supervision into CD to improve performance. \citet{liu2020building} and \citet{zheng2022mdesnet} proposed simultaneously addressing the CD task and semantic segmentation task. \citet{zhang2020deeply} employed deep supervision to ensure the network was well-trained.

Overall, the relation between changed pixels and the relationship between changed pixels and unchanged pixels cannot be controlled directly with these methods. Our method is different in the following aspects: First, we learn the semantics of changes in the difference feature space instead of only learning the semantics of objects in bi-temporal images. Second, we force various changed pixels to share the same common changes' knowledge, thereby controlling relationships between pixels more directly.
\begin{figure*}[!t]
	\centering
	\includegraphics[width=1\linewidth]{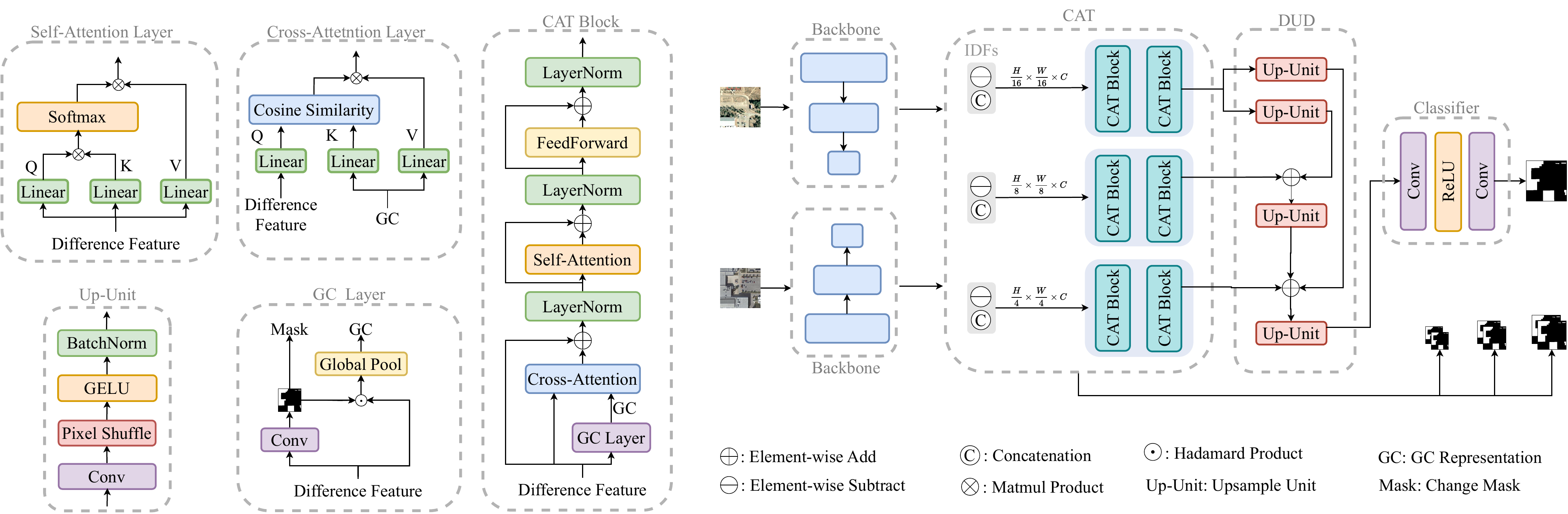}
	\caption{Architecture of CAT and our CD model. Zoom in for a better view.}
	\label{fig_overview}
\end{figure*} 
\subsection{Transformer}
Attention mechanism and Transformer \cite{bahdanau2015neural, vaswani2017attention} has shown great applicability in computer vison (CV) tasks like image classification \cite{dosovitskiy2020image, liu2021swin, wang2021pyramid}, semantic segmentation \cite{zheng2021rethinking, xie2021segformer, strudel2021segmenter}, object detection \cite{carion2020end, chen2023ddod, yuan2021temporal, dai2022ao2},  visual tracking \cite{wang2021transformer, jiao2021deep, wang2022mfgnet, tang2023learning}, and also CD \cite{chen2021remote, bandara2022transformer, zhang2022swinsunet}. There are two main types of Transformers in CV: global attention-based Transformers \cite{dosovitskiy2020image, wang2021pyramid} and window attention-based Transformers \cite{liu2021swin, zhang2022nested}. In CD, BIT \cite{chen2021remote} and Changeformer \cite{bandara2022transformer} are built based on global Attention Transformers. BIT tokenized bi-temporal image features extracted from the ResNet18 \cite{he2016deep} and concatenated them into one token sequence. Then, BIT fed them into the Transformer similar to ViT \cite{dosovitskiy2020image}, and aimed at learning the spatial-temporal global dependence across the image pairs to enhance the image representations. The difference feature map is the absolute value of the subtraction of the two enhanced features. ChangeFormer proved that using Transformer as a backbone can achieve great performance. They employed a mini-convolution network to fuse image features to obtai the difference features. SwinSUNet \cite{zhang2022swinsunet} constructed a network similar to FC-Siam-conc \cite{daudt2018fully} using SwinTransformer to explore the potential of a pure Transformer framework. Unlike these Transformer-based CD models, our network employs cosine cross-attention to perceive changed pixels and update the difference features in an explicit way. And it performs pixel-wise local window-based self-attention.

\section{Changes-Aware Transformer}
In this section, we elaborate on the CAT block and the model integrating CAT to address CD. Fig. \ref{fig_overview} shows the architecture of CAT and the CD model integrating it. A basic block of CAT consists of the GC representation learning layer (Sec. \ref{sec_gc}), the cosine cross-attention layer (Sec. \ref{sec_crossattn}), the local window self-attention layer (Sec. \ref{sec_selfattn}), and the feedforward network (Sec. \ref{sec_selfattn}). Multiple blocks stack to construct an independent module of CAT for processing an IDF of a resolution. In our deployment, each module has 2 blocks. Furthermore, we develop a local window attention-based Transformer backbone and a dense upsampling module for CD (Sec. \ref{sec_cdmodel}). The backbone network produces bi-temporal image features. The difference feature maps outputted by CAT are sent to our dense upsampling module for upsampling and multi-scale feature fusion. Finally, the classifier takes the fused difference features as an input for binary classification.

\subsection{GC Representation Learning Layer}\label{sec_gc} 
The CAT works in the difference feature space, and we combine learning-based and subtraction-based methods to obtain the initial difference features. Specifically, we employ a convolution layer with a kernel size of $ 3\times 3 $ to fuse the bi-temporal features concatenated along the channel direction. Then, we add the difference between the post-temporal and the pre-temporal features to the convolved feature. Let $ \mathbf{X}^{i}_{1} \in \mathbb{R}^{H^{i}\times W^{i}\times C^{i}} $ and $ \mathbf{X}^{i}_{2} \in \mathbb{R}^{H^{i}\times W^{i}\times C^{i}} $ be the bi-temporal image features of the $ i $-th ($ i\in [1, 2, 3] $ in our model) resolution, $ H^{i} $, $ W^{i} $, and $ C^{i} $ be the height, width, and number of channels of the images features, respectively, the IDF $ \mathbf{X}^{i}_{IDF} \in \mathbb{R}^{H^{i}\times W^{i}\times C^{i}} $ is:	
\begin{equation}
	\mathbf{X}^{i}_{IDF} = Conv_{3\times3}(Concat(\mathbf{X}^{i}_{1}, \mathbf{X}^{i}_{2})) + |\mathbf{X}^{i}_{2} -\mathbf{X}^{i}_{1}|,
\end{equation}
where $ Conv_{3\times3} $ means the convolution layer, and $ Concat $ means the concatenation operation.

To make the GC representation learn common knowledge of changes from the difference features, we need some changed pixels belonging to different kinds of changes as raw materials. We utilize existing difference features and deep supervision \cite{lee2015deeply} to learn a mask that can filter changed pixels for us. It is equivalent to predicting changed pixels using non-refined difference features. Note that this mask cannot identify all changed pixels, but it can most likely find some pixels of each change. Concretely, we use a single convolution layer with a kernel size of $ 3\times 3 $ to predict the change mask under the guidance of deep supervision in training. By elements-wise multiplying this change mask and the input difference features, we obtain a part of the changed pixels in the difference feature space. Then, we take the average of these pixels as the GC representation, which means the GC representation is the linear combination of changed pixels. In our implementation, we leverage global pooling to realize this average. Ideally, the denominator in the average operation should be the number of filtered changed pixels, while the number of all pixels is used in the global pooling. However, since the denominator is constant, it does not affect the cosine similarity measure we use later.

Denote the input difference features of this layer as $ \mathbf{X}_{in} \in \mathbb{R}^{H \times W \times C} $, where $ H, W, $ and $ C $ are the height, width, and number of feature channels, respectively. Formulaically, the calculation of the GC representation $ \mathbf{x}_{gc} \in \mathbb{R}^{1\times C}$ is represented as:
\begin{align}
	&Mask = \boldsymbol{\sigma^1}(Conv(\mathbf{X}_{in})) \\
	&\mathbf{x}_{gc} = GlobalAVGPool2d(\mathbf{X}_{in} \odot Mask),
\end{align}
where $ \odot $ is element-wise production, $ \boldsymbol{\sigma} $ is the softmax function, and the $ \boldsymbol{\sigma^1} $ refers to use the channel that indicate the probability of a pixel is a changed pixel.

\subsection{Cross-attention Layer}\label{sec_crossattn}
Given the GC representation containing various kinds of change knowledge, we hope to embed it into changed pixels to enhance their distinctiveness. Inspired by the query-key mechanism, we adopt the cross-attention layer to perceive changed pixels and update the difference features. Concretely, we employ all the pixels as the queries and the GC representation as the key and value. The GC representation explicitly offers an awareness of various changed pixels by measuring the correlation between it (key) and each pixel (query). Because the GC representation is the linear combination of changed pixels in the difference feature space, we measure the linear correlation and adopt cosine similarity as the measurement function. The greater the absolute value of the cosine similarity between a pixel and the GC representation, the more likely it is to be a changed pixel. Then, pixels' difference features are updated under the guidance of these similarities, i.e., cross-attention weights. In this way, the changed pixels in the difference feature space are enhanced by the GC representation, while unchanged pixels irrelated to the GC representation are not enhanced by it. As a result, the difference features are more distinctive.

The output of cross-attention calculation is wrapped by residual connection \cite{he2016deep} and Layer Normalization \cite{ba2016layer}. We follow \cite{vaswani2017attention} to construct the multi-head cosine cross-attention to obtain multiple representation subspaces. Let $ \mathbf{X}_{df} \in \mathbb{R}^{N \times C} $ be the input difference features, $ \mathbf{\hat{X}}_{df} \in \mathbb{R}^{N \times C} $ be the enhanced difference features, $ N= H\times W $, the projection weights and biases corresponding to queries $ \mathbf{Q} \in \mathbb{R}^{N\times C} $, key $ \mathbf{k} \in \mathbb{R}^{1\times C} $, and value $ \mathbf{v} \in \mathbb{R}^{1\times C}$ be $ \mathbf{W}_Q \in \mathbb{R}^{C\times C},\mathbf{b}_Q\in \mathbb{R}^{C\times 1} $, $ \mathbf{W}_k \in \mathbb{R}^{C \times C},\mathbf{b}_k\in \mathbb{R}^{C\times 1} $, and $ \mathbf{W}_v \in \mathbb{R}^{1\times C}, \mathbf{b}_v\in \mathbb{R}^{C\times 1} $, $ \mathbf{1} \in \mathbb{R}^{N\times 1} $ be a column vector with all the elements being $ 1 $, and $ \mathbf{Q}_{row} = [||\mathbf{q}^{\top}_1||_2; ||\mathbf{q}^{\top}_2||_2;\dots;||\mathbf{q}^{\top}_N||_2] $,
the cross-attention layer is expressed in formulas as follows:

\begin{align}
	\mathbf{Q} &= \mathbf{X}_{df}\mathbf{W}_Q+\mathbf{1}\mathbf{b}_Q^\top,\\
	\mathbf{k} &= \mathbf{x}_{gc}\mathbf{W}_k+\mathbf{b}_k^\top,\\ 
	\mathbf{v} &= \mathbf{x}_{gc}\mathbf{W}_v+\mathbf{b}_v^\top
\end{align}
\begin{align}
	&\mathbf{A}_{cross} = \dfrac{\mathbf{Q} \mathbf{k}^{\top}}{\mathbf{Q}_{row}||\mathbf{k}||_2} \mathbf{v},\\  &\mathbf{\hat{X}}_{df} = LayerNorm(\mathbf{X}_{df} + \mathbf{A}_{cross})
\end{align}

\subsection{Local Window Self-attention Layer}\label{sec_selfattn}
We apply the local window self-attention layer following the cosine cross-layer to learn the context of pixels in the difference feature space. The basic unit in NesT makes up this layer, and please refer to \cite{zhang2022nested} for more details. Here, we elaborate on the differences between our self-attention layer and NesT's.
First, the non-overlap window size is $ 8\times 8 $ in CAT, whereas that in Nest is $ 14\times 14 $. In this way, the number of windows for each resolution feature in CAT is $ 64 $, $ 16 $, and $ 4 $. As a result, none of our self-attention layers has a global receptive field, while NesT performs global self-attention on the entire feature map at its top level. Our observation is that the interested changed pixels are always spatially aggregated, e.g., buildings. Therefore, we assume whether a pixel is changed can gain enough auxiliary information from a large neighbor region, which is not necessarily the whole picture. In our model, this large receptive field's size is 1/4 of the image, which is $ 128\times 128 $. Secondly, we delete the position encoding because position information does not contain the general characteristics of changes (changes can occur anywhere in an image) and may bring noise to the learned generalized knowledge. 

The architecture of the feedforward network following the self-attention layer is the same with the classical Transformers \cite{vaswani2017attention,wang2021pyramid,zhang2022nested}.

\subsection{Integrate CAT to Change Detection Model}\label{sec_cdmodel}
\subsubsection{Backbone}
CAT can be integrated into Transformer-based and CNN-based backbone networks and also existing methods. Specifically, we employ the small version of NesT as the Transformer-based Siamese \cite{chopra2005learning} backbone and make three modifications to better adapt to CD. First, we utilize an input image with a size of $ 256\times 256 $, while the input size in the original network is $ 224\times 224 $. Secondly, same as the local self-attention layer in CAT, all the self-attention operations are conducted in local windows. Third, we stack more encoder layers at the first two stages because of their feature maps have larger spatial sizes. Also, we reduce the number of layers at the last stage compared to the NesT. As a result, we have a 4-4-6 for layers setting while NesT's is 2-2-20. This modification provides more detailed image information. With these changes, our Transformer-based backbone works efficiently in CD and has better performance. Additionaly, we deploy a model with ResNet18 \cite{he2016deep} as the Siamese backbone network and extract multi-scale features from the conv2, conv3, and conv4 layers. When integrates CAT to prior work ChangeFormer, we replace the Difference Module in ChangeFormer by our CAT while keep it's retaining modules.

\subsubsection{Dense Upsample Decoder}
Change Detection is a task that requires pixel-wise detailed information. Hence, we advocate that as much information as possible in all downsampled features should be restored to high-resolution feature maps. The dense upsample decoder (DUD) to fuse and retain information from the multi-scale difference feature maps thoroughly is developed in our work. In detail, we utilize Pixcel Shuffle \cite{shi2016real} as the basic upsample method, and the upsampling unit in DUD consists of a convolution layer with a kernel size of $ 1\times 1 $, a pixel shuffle operation layer, a GELU activation layer \cite{hendrycks2016gaussian}, and a Batch Normalization layer\cite{ioffe2015batch}. Each resolution difference feature is upsampled to all the resolutions greater than it by a corresponding upsample unit, and the new difference feature is the sum of the original one and the upsampling one. 

The upsample scheme starts from the difference feature with the smallest spatial resolution $ 16\times 16 $. Two separate upsample units upsample it to resolutions of $ 32\times 32 $ and $ 64\times 64 $, respectively. Then, the old difference features with the resolution of $ 32\times 32 $ and $ 64\times 64 $ add the two upsampling features to form the new difference features. Next, DUD performs the upsampling of the $ 32\times 32 $ spatial resolution feature. Finally, the difference feature with the resolution of $ 64\times 64 $ is upsampled to the original images' size ($ 256 \times 256 $).

\begin{table}
	\caption{Results on LEVIR. "Pre." means Precision. "Rec." means Recall.}
	\label{tab_levir}
	\centering
	\footnotesize
	\begin{tabular}{@{}ccccc@{}}
		\toprule
		Model        & Backbone    & F1    & Pre. & Rec.    \\ \midrule
		FC-EF\cite{daudt2018fully}        & FCNN\cite{daudt2018fully} & 83.40 & 86.91     & 80.17 \\
		FC-Siam-Conc\cite{daudt2018fully} & FCNN\cite{daudt2018fully} & 83.69 & 91.99     & 76.77 \\
		FC-Siam-Di\cite{daudt2018fully}   & FCNN\cite{daudt2018fully} & 86.31 & 89.53     & 83.31 \\
		IFNet\cite{zhang2020deeply}        & VGG16\cite{simonyan2014very} & 88.13 & \textbf{94.02}     & 82.93 \\
		DDCNN\cite{peng2020optical}        & UNet++\cite{peng2019end} & 90.24 & 91.85     & 88.69 \\
		DESSN\cite{lei2021difference}        & UNet\cite{ronneberger2015u} & 91.36 & 90.99     & \textbf{91.73} \\
		ST-MSA\cite{zhou2022spatial}        & ResNet18\cite{he2016deep} & 88.20 & 86.90     & 89.50 \\
		BIT\cite{chen2021remote}          & ResNet18\cite{he2016deep} & 89.31 & 89.24 & 89.37 \\ \midrule
		ChangeFormer\cite{bandara2022transformer} & PVT\cite{wang2021pyramid} & 89.96 & 91.83     & 88.17 \\
		CAChangeFormer\cite{bandara2022transformer} & PVT\cite{wang2021pyramid} & 91.12 & 92.85     & 89.46 \\ \midrule
		ResNet-Siam          & ResNet18\cite{he2016deep} & 89.95 & 91.53 & 88.44 \\
		\textit{CAT-Siam-R}          & ResNet18\cite{he2016deep} & 90.39 & 91.30 & 89.50 \\ \midrule
		NesT-Siam          & NesT & 90.46 & 92.66 & 88.36 \\
		\textit{CAT-Siam-T}          & LocalNesT & \textbf{91.53} & 93.29 & 89.84 \\ \bottomrule
	\end{tabular}
\end{table} 
\begin{table}
	\caption{Results on DSIFN. "Pre." means Precision. "Rec." means Recall.}
	\label{tab_dsifn}
	\centering
	\footnotesize
	\begin{tabular}{@{}ccccc@{}}
		\toprule
		Model        & Backbone    & F1    & Pre. & Rec.    \\ \midrule
		FC-EF\cite{daudt2018fully}        & FCNN\cite{daudt2018fully} & 61.09 & 72.61     & 52.73 \\
		FC-Siam-Conc\cite{daudt2018fully} & FCNN\cite{daudt2018fully} & 59.71 & 66.45     & 54.21 \\
		FC-Siam-Di\cite{daudt2018fully}   & FCNN\cite{daudt2018fully} & 62.54 & 59.67     & 65.71 \\
		IFNet\cite{zhang2020deeply}        & VGG16\cite{simonyan2014very} & 60.10 & 67.86     & 53.94 \\
		ST-MSA\cite{zhou2022spatial}        & ResNet18\cite{he2016deep} & 65.90 & 59.00     & 74.70 \\
		BIT\cite{chen2021remote}          & ResNet18\cite{he2016deep} & 75.13 & 88.76 & 65.13 \\ \midrule
		ChangeFormer\cite{bandara2022transformer} & PVT\cite{wang2021pyramid} & 86.67 & 88.48     & 84.94 \\
		CAChangeFormer\cite{bandara2022transformer} & PVT\cite{wang2021pyramid} & 94.15 & 94.04     & 94.26 \\ \midrule
		ResNet-Siam          & ResNet18\cite{he2016deep} & 75.91 & 66.99 & 87.58 \\
		\textit{CAT-Siam-R}          & ResNet18\cite{he2016deep} & 87.15 & 85.84 & 88.50 \\ \midrule
		NesT-Siam          & NesT & 92.72 & 93.13 & 92.32 \\
		\textit{CAT-Siam-T}          & LocalNesT & \textbf{94.81} & \textbf{94.87} & \textbf{94.75} \\ \bottomrule
	\end{tabular}
\end{table}

\subsubsection{Classfification Network and Loss Function}
According to the final upsampled feature maps, a binary classification network determines whether or not a pixel has changed. The whole framework is trained end-to-end. We employ Cross-Entropy Loss as the cost function for the GC representation learning layers and the final prediction. Denote the final change map as $ \mathbf{X}_{out} \in \mathbb{R}^{256\times 256 \times 2} $, the change mask of $ j $-th block in the $ i $-th module of CAT as $ Mask^{ij} \in \mathbb{R}^{H^i \times W^i \times 2} $, $ i = [1, 2, 3] $, $ j = [1, 2] $, hyper-parameters $ \lambda_{ij} = 1 $, the full loss is:
\begin{equation}
	\begin{aligned}
		Loss =&\ CE(\mathbf{X}_{out}, label) \\
		&+ \sum_{i}\sum_{j} \lambda_{ij}CE(Mask^{ij}, label^{i})
	\end{aligned}
\end{equation}

\section{Experiments and Analysis}
\subsection{Data Sets and Evaluation Metrics}
To evaluate performance of the proposed CAT, we conducted experements on three remote sensing CD data sets and one street scene CD data set, i.e., LEVIR \cite{chen2020spatial}, DSIFN \cite{zhang2020deeply}, CDD \cite{lebedev2018change}, and VL-CMU-CD\cite{alcantarilla2018street}. 

The \textbf{LEVIR} data set focuses on building change detection and contains $ 637 $ image pairs with a size of $ 1024\times 1024 $ and a resolution of $ 0.5 $m. It covers $ 20 $ different regions and time intervals of $ 5 $ to $ 14 $ years. Noisy elements like season changes and illumination changes are included. We follow \cite{bandara2022transformer} to cut the images to non-overlap patches with a size of $ 256\times 256 $ and split them into training, validation, and test subsets. The number of image patches for each subset is $ 7120 $, $ 1024 $, and $ 2048 $, respectively. 

The \textbf{DSIFN} data set consists of image pairs with a size of $ 512\times 512 $ and a resolution of $ 2 $ m. It covers the changes in city ground objects of six large cities in China. Following the data process of \cite{bandara2022transformer}, we have $ 14400 $, $ 1360 $, and $ 192 $ non-overlap image patch pairs with a size of $ 256\times 256 $ for training, validation, and test, respectively.

The \textbf{CDD} data set has $ 7 $ real image pairs with a size of $ 4725\times 7200 $. It includes natural object changes and manufactural object changes. The publicized default training, validation, and test subset has $ 10000 $, $ 3000 $, and $ 3000 $ samples each. All the samples have a size of $ 256\times 256 $.

For the three data sets above, we followed previous works \cite{lei2021difference,chen2021remote,zhang2022swinsunet} trained our model on the training subset and evaluated it on the validation subset when each training epoch was done. Ultimately, the best validation model was tested on the test subset. We report test results in the later content.

The \textbf{VL-CMU-CD} data set contains changes in street view. The bi-temporal image pairs are extracted from videos captured by a vehicle-mounted monocular camera, including $ 933 $ training samples and $ 429 $ test samples. There are $ 10 $ categories of objects, including barriers, bins, construction maintenance, personcycles, rubbish bins, sign boards, traffic cones, vehicles, and others. Hence, the change detection task can be formed as binary change detection (changed or unchanged) and multi-class change detection (indicate the specific category of the changed object or no change). Following previous works \cite{chen2021dr, wang2023reduce}, we used rotation to augment the training data and obtained $ 3732 $ samples to conduct the binary change detection task, and the results are reported on the test data set.

The prevalent metrics for verifying the performance of a binary classification CD method include Precision, Recall, and F1 score, with regard to the change category. TP refers to the number of true positives, FP refers to the number of false positives, TN is the number of true negatives, and FN refers to the number of false positives, formulaically, these metrics are represented as:

\begin{align}
	&Precision = \dfrac{TP}{TP+FP},\\
	&Recall = \dfrac{TP}{TP+FN},\\
	&F1 = \dfrac{2}{Precision^{-1}+Recall^{-1}}
\end{align}

\subsection{Implement Details}
All our experiments were conducted on 1-2 NVIDIA GeForce RTX 2080Ti GPU using Pytorch \cite{paszke2019pytorch}. The data augmentation techniques we used for the LEVIR, DSIFN, and CDD data sets was the same to \cite{bandara2022transformer}. For the experiments on the VL-CMU-CD data set, we following \cite{wang2023reduce} to augment the training data. The number of training epochs for our models is $ 200 $. AdamW\cite{loshchilov2018decoupled} with the weight decay of $ 0.01 $ is the optimizer, and we employed a linear decay strategy with an initial learning rate of $2e-4$. The total batch size is $16$.

\subsection{Compare with SOTA}
\begin{table}
	\caption{Results on CDD. "Pre." means Precision. "Rec." means Recall.}
	\label{tab_cdd}
	\centering
	\footnotesize
	\begin{tabular}{@{}ccccc@{}}
		\toprule
		Model        & Backbone    & F1    & Pre. & Rec.    \\ \midrule
		FC-EF\cite{daudt2018fully}        & FCNN\cite{daudt2018fully} & 59.80 & 77.58     & 48.66 \\
		FC-Siam-Conc\cite{daudt2018fully} & FCNN\cite{daudt2018fully} & 69.03 & 81.00     & 60.14 \\
		FC-Siam-Di\cite{daudt2018fully}   & FCNN\cite{daudt2018fully} & 69.07 & 78.14     & 61.88 \\
		IFNet\cite{zhang2020deeply}          & VGG16\cite{simonyan2014very} & 90.30 & 94.96 & 86.08 \\
		DDCNN\cite{peng2020optical}        & UNet++\cite{peng2019end} & 94.46 & 96.71     & 92.32 \\
		DESSN\cite{lei2021difference}        & UNet\cite{ronneberger2015u} & 91.80 & 95.04     & 88.77 \\
		ESCNet\cite{zhang2021escnet}       & UNet\cite{ronneberger2015u} & 93.51 & 90.04     & 97.26 \\
		BIT\cite{chen2021remote}          & ResNet18\cite{he2016deep} & 94.77 & 94.82 & 94.72 \\
		SwinSUNet\cite{zhang2022swinsunet}    & SwinT\cite{liu2021swin} & 94.00 & 95.70     & 92.30 \\ \midrule
		ChangeFormer\cite{bandara2022transformer} & PVT\cite{wang2021pyramid} & 94.34 & 94.52 & 94.15 \\
		CAChangeFormer\cite{bandara2022transformer} & PVT\cite{wang2021pyramid} & \textbf{97.75} & \textbf{97.36}     & \textbf{98.15} \\ \midrule
		ResNet-Siam          & ResNet18\cite{he2016deep} & 94.89 & 95.57 & 94.21 \\
		\textit{CAT-Siam-R}          & ResNet18\cite{he2016deep} & 96.13 & 96.24 & 96.02 \\ \midrule
		NesT-Siam          & NesT & 96.95 & 96.78 & 97.11 \\
		\textit{CAT-Siam-T}          & LocalNesT & 97.49 & 97.30 & 97.68 \\ \bottomrule
	\end{tabular}
\end{table}

\begin{table}
	\caption{Results on VL-CMU-CD. "Pre." means Precision. "Rec." means Recall. "N/A$ ^{\dagger} $" means the traning of FC-EF on this data set did not converge.}
	\label{tab_cmu}
	\centering
	\footnotesize
	\begin{tabular}{@{}ccccc@{}}
		\toprule
		Model        & Backbone    & F1    & Pre. & Rec.    \\ \midrule
		FC-EF\cite{daudt2018fully}        & FCNN\cite{daudt2018fully} & N/A$ ^{\dagger} $ & N/A$ ^{\dagger} $    & N/A$ ^{\dagger} $ \\
		FC-Siam-Conc\cite{daudt2018fully} & FCNN\cite{daudt2018fully} & 52.57 & 79.33     & 39.31 \\
		FC-Siam-Di\cite{daudt2018fully}   & FCNN\cite{daudt2018fully} & 41.76 & \textbf{82.47}     & 27.96 \\
		DR-TANet\cite{zhang2020deeply}          & ResNet18\cite{he2016deep} & 57.98 & 68.46 & 50.29 \\
		C-3PO\cite{peng2020optical}        & ResNet18\cite{he2016deep} & 65.06 & 70.09     & 60.70 \\ \midrule
		ResNet-Siam          & ResNet18\cite{he2016deep} & 57.90 & 64.86 & 52.28 \\
		\textit{CAT-Siam-R}          & ResNet18\cite{he2016deep} & \textbf{69.35} & 74.73 & \textbf{64.69} \\ \midrule
		NesT-Siam          & NesT & 51.55 & 74.42 & 39.43 \\
		\textit{CAT-Siam-T}          & LocalNesT & 64.69 & 75.94 & 56.34 \\ \bottomrule
	\end{tabular}
\end{table}

We named our model CAT-Siam-T, it integrates the modified local NesT backbone, CAT, DUD, and the classifier. Replace the local NesT backbone with ResNet18, we have our CAT-Siam-R model. The baseline networks are the original NesT with 14 layers and the ResNet18 with its conv2, conv3, and conv4 layers. We named the former NesT-Siam and the later ResNet-Siam. The upsampling module is DUD for the two baselines. We also use ChangeFormer to be a baseline model and evaluate CAT's efficacy when integrate it into an existing method. The integrated model is CAChangeFormer. We compare them with SOTA methods to demonstrate our models' effectiveness. These methods include FC-EF \cite{daudt2018fully}, FC-Siam-Di \cite{daudt2018fully}, FC-Siam-Conc \cite{daudt2018fully}, IFNet \cite{zhang2020deeply}, DDCNN \cite{peng2020optical}, DESSN \cite{lei2021difference}, ESCNet \cite{zhang2021escnet}, BIT \cite{chen2021remote}, ChangeFormer \cite{bandara2022transformer}, SwinSUNet \cite{zhang2022swinsunet}, DR-TANet\cite{chen2021dr}, and C-3PO \cite{wang2023reduce}.

\begin{figure*}[!t]
	\centering
	\includegraphics[width=1\linewidth]{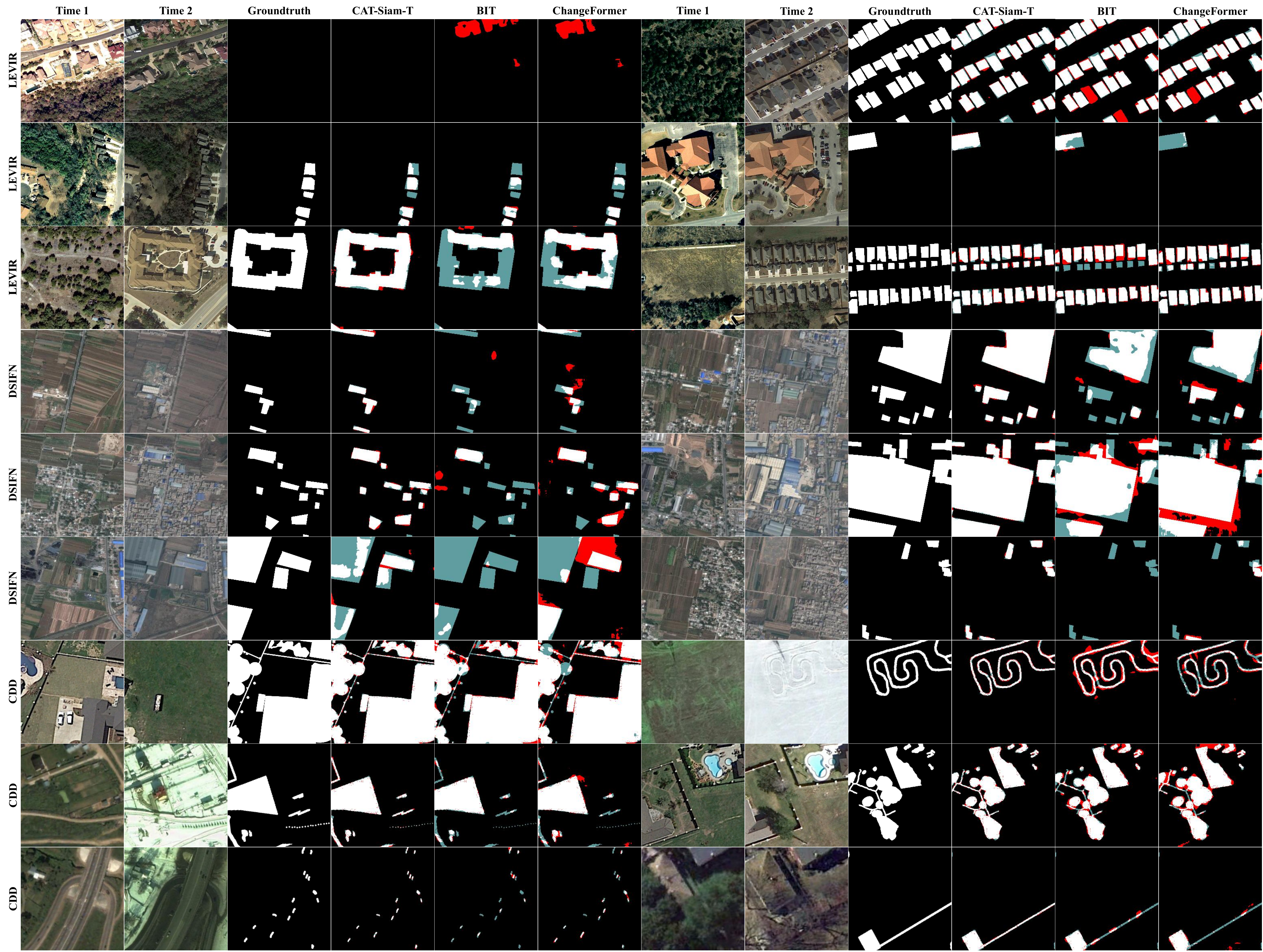}
	\caption{Comparision with SOTA methods on remote sensing CD data sets. White color indicates TP, black indicates TN, turquoise indicates FN, and red indicates FP.}
	\label{fig_sota}
\end{figure*}
\begin{figure*}[!t]
	\centering
	\includegraphics[width=1\linewidth]{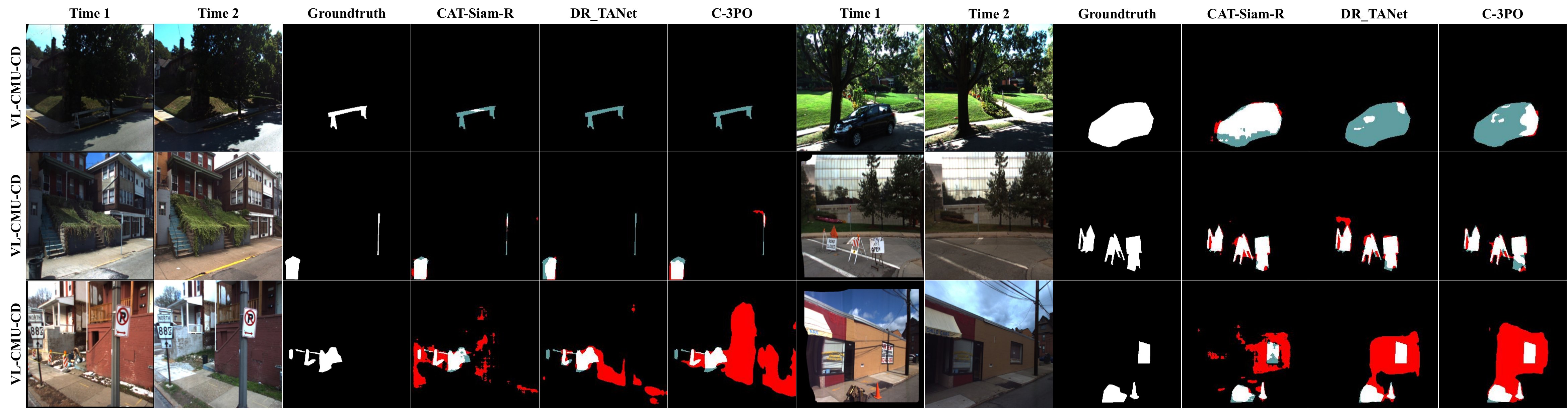}
	\caption{Comparision with SOTA methods on the VL-CMU-CD data set. White color indicates TP, black indicates TN, turquoise indicates FN, and red indicates FP.}
	\label{fig_cmu_sota}
\end{figure*}
\subsubsection{Results}
Tabel \ref{tab_levir}, \ref{tab_dsifn}, and \ref{tab_cdd} list the results on remote sensing change detection data sets. On the three data sets, CAT-Siam-T brings improvements of 1.07/2.09/0.54\% for the F1 score to the baseline NesT-Siam. CAT-Siam-R outperforms the ResNet-Siam by 0.44/11.24/1.24\% for the F1 score. And adding CAT to ChangeFormer brings F1 score improvements of 1.12/7.48/3.41\%. Models equipped with CAT all have higher Recall than baselines, which illustrates that CAT is beneficial for recognizing more changed pixels. Note that CAT gives large benefits to the performances on the DSIFN data set. We attribute this to the GC representation's awareness of different kinds of changed pixels and enhancement of the difference features. Because the expansive view of urban regions covers unified change semantics (mainly the land becomes buildings) but more kinds of changes (various buildings), CAT learns the generalized change knowledge and embeds it into various changed pixels in the difference feature space.

Compared with SOTA methods, CAT-Siam-T achieves the highest F1 scores on the LEVIR and DSIFN data sets, CAChangeFormer has the highest F1 score on the CDD data set, and CAT-Siam-R produce leading F1 scores on the DSIFN and CDD and competitive result on the LEVIR. For the Transformer-based approaches, CAT-Siam-T exceeds BIT by 2.22/4.05/0.47\% and exceeds ChangeFormer by 1.57/1.46/1.67\% for F1/Precision/Recall on the LEVIR. On the DSIFN and CDD, CAT-Siam-T and CAChangeFormer improve the performance by a large margin. For instance, CAT-Siam-T exceeds ChangeFormer by 8.14/6.39/9.81\% for F1/Precision/Recall on the DSIFN and exceeds SwinSUNet by 3.49/1.6/5.38\% on the CDD. Compared with the CNN-based model DESSN on the LEVIR, CAT-Siam-R and CAChangeFormer's performance are lower, CAT-Siam-T has a higher F1 score and Precision, lower Recall. However, our three models outperform DESSN by a large margin on the CDD, for instance, CAT-Siam-T exceeds 5.7/2.3/8.9\% for F1/Precision/recall. Overall, our models have better generalization and address different types of changes well. 

A variety of very different street scenes and less data make the VL-CMU-CD data set quite challenging. The experimental results are listed in Table \ref{tab_cmu}. CAT-Siam-R exceeds the ResNet-Siam baseline by 11.45/9.87/12.41\% for F1/Precision/Recall and obtains the highest F1 score. The NesT-Siam baseline has inferior performance compared with CAT-Siam-T, i.e., the 13.14/1.52/16.91\% drop for F1/Precision/Recall. Note that the small margin on the Precision is because of the large margin on Recall, which indicates that the CAT is effective in identifying changed pixels. In addition, we found that CAT-Siam-T and CAT-Siam-R appeared to be overfitting during the training. After training the 30th epoch, the performance of CAT-Siam-T on the training subset continued to improve, but the performance on the test subset did not enhance anymore. CAT-Siam-R's F1 score on the test subset did not increase after the 100th epoch. Compared with other methods, our models are outstanding.

\subsubsection{Visualization} 
We visualize the change maps predicted by BIT, ChangeFormer, and CAT-Siam-T in Fig. \ref{fig_sota}. The LEVIR bi-temporal images on the left of the first row have very different illumination and are easy to cause confusion. In this case, CAT-Siam-T has no FP pixels. The plots of the second row show unapparent change cases. For the sample on the left, the buildings and trees in the post-time image have similar appearances making the changed and unchanged pixels hard to discriminate, and all three methods can not find out all the changed pixels. However, CAT-Siam-T identifies the most. For the sample on the right, the pre-time and post-time images have similar abstract semantics, and the change is an extension of an existing building with an ambiguous edge of the junction. CAT-Siam-T identifies both the changed pixels in the shadow area and the non-shadow area. The third-row samples show CAT-Siam-T can identify large-scale change objects and dense and small-scale change objects.

The DSIFN samples show general manufactural urban change cases and mainly require models to be able to distinguish changes in the cluttered context, including subtle small-scale changes and abstract large-area changes. CAT-Siam-T meets this requirement the best.

For the CDD data set, we find that CAT has a stronger ability to identify finer object changes. We attribute this to two reasons: (1) Changes in CDD's many samples are related to the snow, which provides a clear background and makes the changes simple (e.g., the right sample in the CDD's first row). Our model and GC representation can easily capture these simple changes. (2) Our local attention-based network may have better capability on modeling structure features since local attention works more like a CNN.

Fig. \ref{fig_cmu_sota} illustrates some prediction examples of CAT-Siam-R, DR-TANet, and C-3PO on the VL-CMU-CD data set. The samples on the first row depict the heavy shadow degrades the performance of all three methods, and the CAT-Siam-R recognizes the most number of the changed pixels. Also, CAT-Siam-R exhibits a better ability to recognize fine/small changes, see the second-row figures. The third-row samples show cluttered street scenes, and many FPs are predicted by all three methods. These FPs may caused by the data distribution. For instance, the warning line (left Time 1 image) and the wires (right Time 1 figure) are highly similar, fence changes also appear in other data. Overall, CAT-Siam-R predicts fewer FPs showing better generalization.

\subsection{Ablation Study}
\subsubsection{Analysis of CAT}
We conduct experiments on LEVIR to verify the effectiveness of the components in CAT, see Table \ref{tab_ablation}. The modified NesT Siamese backbone network, DUD, and classification network are retained in all models. 

The configuration of \textbf{w/o\_CAT} removes the CAT from CAT-Siam-T, and its performance drops 0.55/1.09/0.05\% for F1/Precision/Recall. This result demonstrates that CAT-Siam-T has better capability on distinguishing changed and unchanged pixels because it has less FP at almost the same Recall. Also, our local window attention-based Transformer backbone learns useful image representations for CD. 

For the setting of \textbf{w/o\_GC}, we remove the GC representation and the cross-attention layer but retain deep supervision. This model is equivalent to using another local window attention-based Transformer to refine difference features, and it recognizes fewer changed pixels. The Recall drops by 0.60\%. It indicates fine-tuning of difference features is low efficient if the GC representation-based cross-attention is not used. And the improvements brought by CAT do not stem from deep supervision. 

The \textbf{w/o\_SA} model removes the self-attention layer from CAT-Siam-T's CAT module but preserves the GC representation and cross-attention layer. It shows less performance degradation, suggesting that the cosine cross-attention layer with the GC representation contributes the most to CAT's performance.

\begin{table}
	\caption{Performance comparisions of CAT. "DS" means deep supervision, "MCA" means multi-head cross-attention layer, "MSA" means multi-head self-attention layer. "Pre." means Precision. "Rec." means Recall.}
	\label{tab_ablation}
	\centering
	\footnotesize
	\begin{tabular}{@{}ccccccc@{}}
		\toprule
		Model                       & DS & MCA & MSA & F1    & Pre.  & Rec.  \\ \midrule
		w/o\_CAT                    & \XSolidBrush  & \XSolidBrush   & \XSolidBrush   & 90.98 & 92.20 & 89.79 \\
		w/o\_GC & \Checkmark  & \XSolidBrush   & \Checkmark   & 91.05 & 92.94 & 89.24 \\
		w/o\_SA & \Checkmark  & \Checkmark   & \XSolidBrush   & 91.31  & 93.04 & 89.65  \\
		CAT-Siam-T                  & \Checkmark  & \Checkmark   & \Checkmark   & \textbf{91.53} & \textbf{93.29} & \textbf{89.84} \\ \bottomrule
	\end{tabular}
\end{table}

\subsubsection{Visualization}
\begin{figure*}[!t]
	\centering
	\subfloat[]{\includegraphics[width=0.49\linewidth]{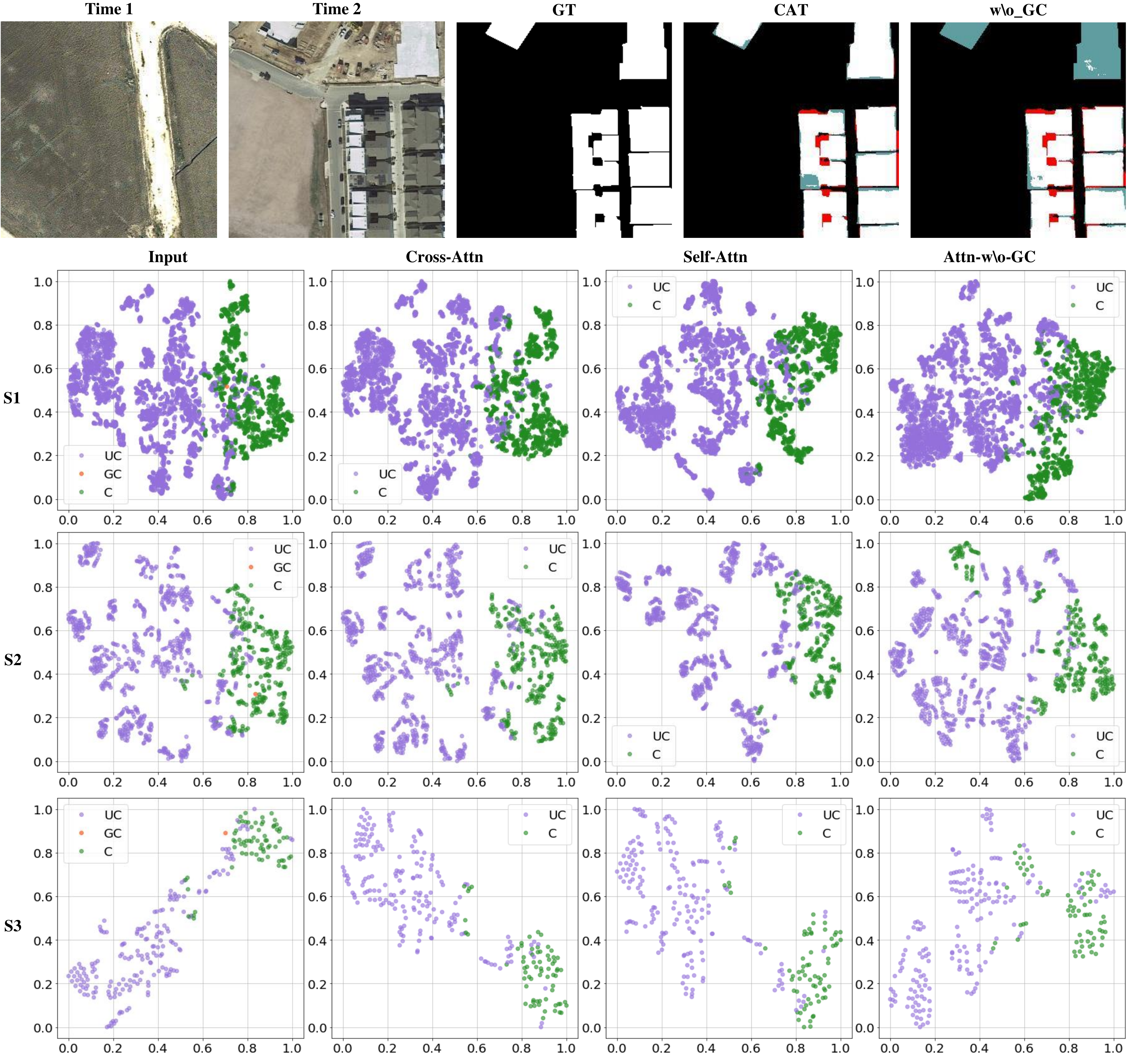}%
		\label{fig_tsne_1}}
	\hfil
	\subfloat[]{\includegraphics[width=0.49\linewidth]{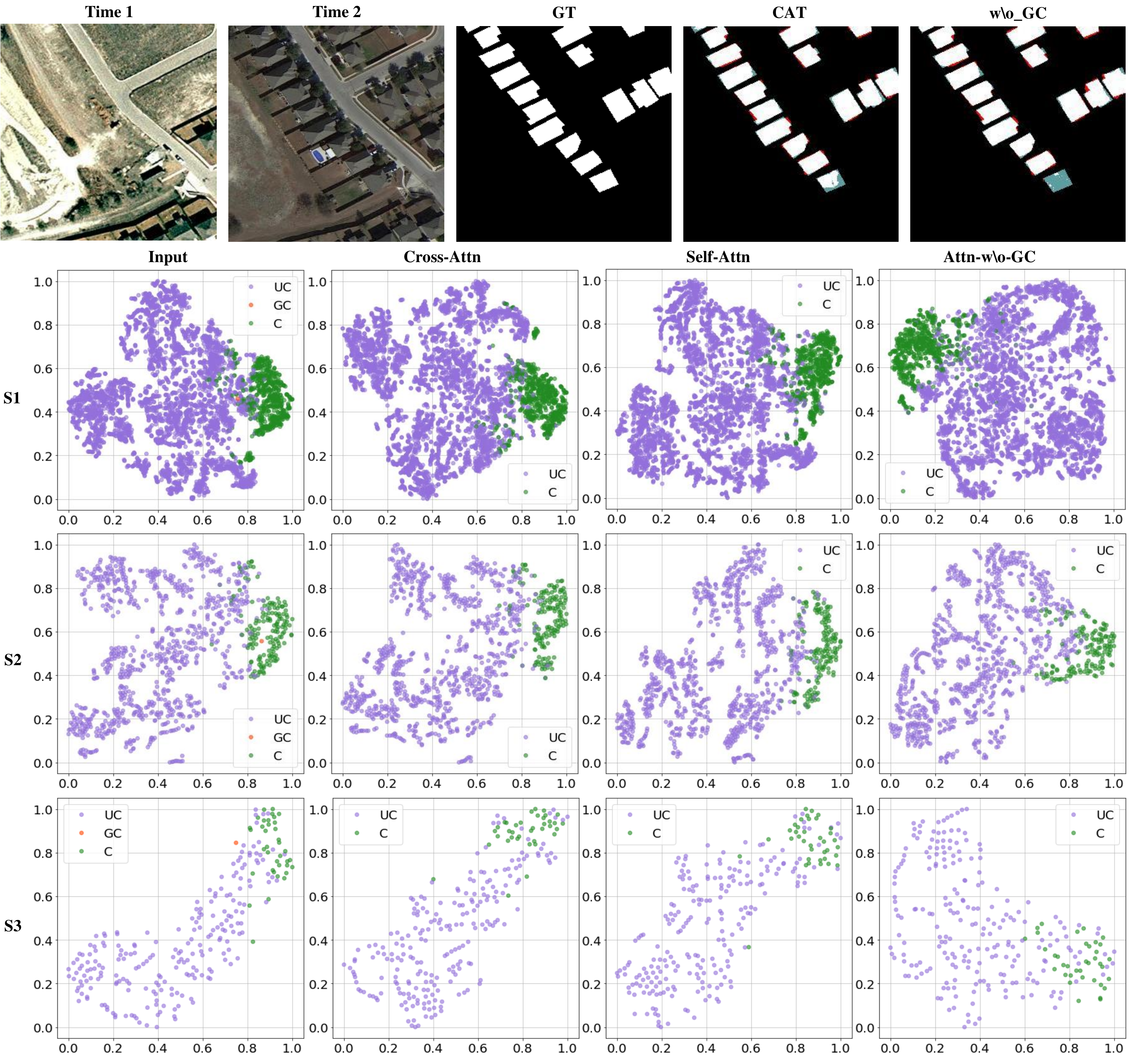}%
		\label{fig_tsne_2}}
	\caption{Visualization of pixels in CAT and w/o\_GC setting. "UC" stands for unchanged pixel. "C" stands for changed pixel. "GC" stands for the GC representation. "Input", "Cross-Attn", and "Self-Attn" stand for the input of the CAT blocks, cosine cross-attention layer, and self-attention layer in CAT, respectively. "Attn-w/o-GC" stands for the self-attention layer in w/o\_GC model. "s1", "s2", and "s3" refer to stage 1, stage 2, and stage 3, which correspond to the features of three spatial resolutions, i.e., $ 64\times 64 $, $ 32\times 32 $, and $ 16\times 16 $.}
	\label{fig_ablation}
\end{figure*}

To better understand the GC representation-based cross-attention layer, we visualize intermediate difference features of CAT-Siam-T and w/o\_GC model in Fig. \ref{fig_ablation}. These features are extracted from the second block of each independent difference feature refinement module of the two models.
Concretely, the difference features input to the GC representation learning layers, the features output by the cosine-cross attention layers, and the features output by the self-attention layers in CAT are extracted. Correspondingly, difference features are extracted from the self-attention layers of w/o\_GC. The resolution of each scale feature is $ 64\times 64 $, $ 32\times 32 $, and $ 16\times 16 $, corresponding to $ 4096 $, $ 1024 $, and $ 256 $ pixels, respectively. For each pixel, we leverage t-SNE \cite{van2008visualizing} to reduce its dimension of channels to $ 2 $ and visualize them in 2D space.

\begin{table}[]
	\caption{Study our model modifications for CD on the LEVIR data set.
		"DUD" means the dense upsample decoder. "Global" means global window. "PE" means position
		encoding.}
	\label{tab_ablation2}
	\centering
	\footnotesize
	\begin{tabular}{@{}ccccccc@{}}
		\toprule
		Model                   & DUD & Global & PE & F1    & Pre.  & Rec.  \\ \midrule
		w/o\_DUD                & \XSolidBrush   & \XSolidBrush        & \XSolidBrush  & 91.23 & 92.51 & 89.98 \\
		w\_Global               & \Checkmark   & \Checkmark        & \XSolidBrush  & 91.19  & 92.49   & 89.93   \\
		w\_PE                   & \Checkmark   & \XSolidBrush        & \Checkmark  & 91.16   & 92.26   & 90.08   \\
		$\lambda=0.5$                  & \Checkmark   & \Checkmark        & \Checkmark  & \textbf{91.53}   & 92.61   & \textbf{90.48}   \\
		$\lambda=0.8$                  & \Checkmark   & \Checkmark        & \Checkmark  & 91.44   & 92.49   & 90.41   \\
		\textit{CAT-Siam-T}              & \Checkmark   & \XSolidBrush        & \XSolidBrush  & \textbf{91.53} & \textbf{93.29} & 89.84   \\ \bottomrule
	\end{tabular}
\end{table}

The plots in the first column of Fig. \ref{fig_tsne_1} and \ref{fig_tsne_2} show that the GC representation is located in/near a cluster of changed pixels. The cosine cross-attention layer tends to gather changed pixels, as shown in pictures of stage 1, changed pixels of the cross-attention layer are closer to each other in the vertical direction than that of the input. Compared with pixels in the pictures of the "Attn\_w/o\_GC" column, the inter-distance of changed pixels in the "Cross-Attn" and "Self-Attn" columns is smaller, especially for stages 3, and the margin between changed pixels cluster and unchanged pixels cluster is larger in the CAT (see Fig. \ref{fig_tsne_1} and \ref{fig_tsne_2}).

\subsubsection{Analysis of Other Modifications}
In Table \ref{tab_ablation2}, we show the influence of our other modified components. Note that the CAT is retained for all the models in Table \ref{tab_ablation2}. In the setting of \textbf{w/o\_DUD}, the basic upsampling unit is the same as that of DUD, but the feature at each level is upsampled to the size of its upper level only. Compared with CAT-Siam-T, its Recall is a little higher, but Precision is lower and decreases by 0.3\% for the F1 score. This demonstrates that directly passing the highest-level pixels' information to the low-level pixels is beneficial for distinguishing changed and unchanged pixels.

In the setting of \textbf{w\_Global}, the self-attention calculations both in the backbone and CAT's last level are performed on the whole feature map. Hence, the final receptive field of w\_Global is $ 256\times 256 $. We think the reason for the performance degradation of w\_Global is updating pixels' features in a larger window may introduce extra noise, because changes are spatially clustered. For example, it can be seen that changed pixels in many samples of Fig. \ref{fig_sota} within the range of $ 128 \times 128 $ receptive field can interact with similar changed pixels. In extreme cases such as the one on the right of the second row of the LEVIR samples in Fig. \ref{fig_sota}, all the changed pixels are only within the range of $ 128 \times 128 $ in the upper left of the image. Therefore, the local but large receptive field (1/4 of the full image) may be enough for a pixel to collect information from other pixels and related object semantics in the self-attention layers.

The \textbf{w\_PE} model adds position encoding to CAT-Siam-T. The increase in Recall may be due to the position encoding providing spatial co-occurrence information, but the decrease in Precision implies it is not a reliable generalized indication for change detection.

For the loss function, we replace the hyper-parameters $ \lambda_{ij} $ of change masks learning by 0.5 and 0.8, respectively. The results show that compared with $ \lambda_{ij} = 1 $, $ \lambda_{ij} < 1 $ increases the Recall and decreases the Precision, and the F1 score is relatively stable.

\subsection{Failure Cases and Limitaions}
\begin{figure*}[!t]
	\centering
	\subfloat[]{\includegraphics[width=0.49\linewidth]{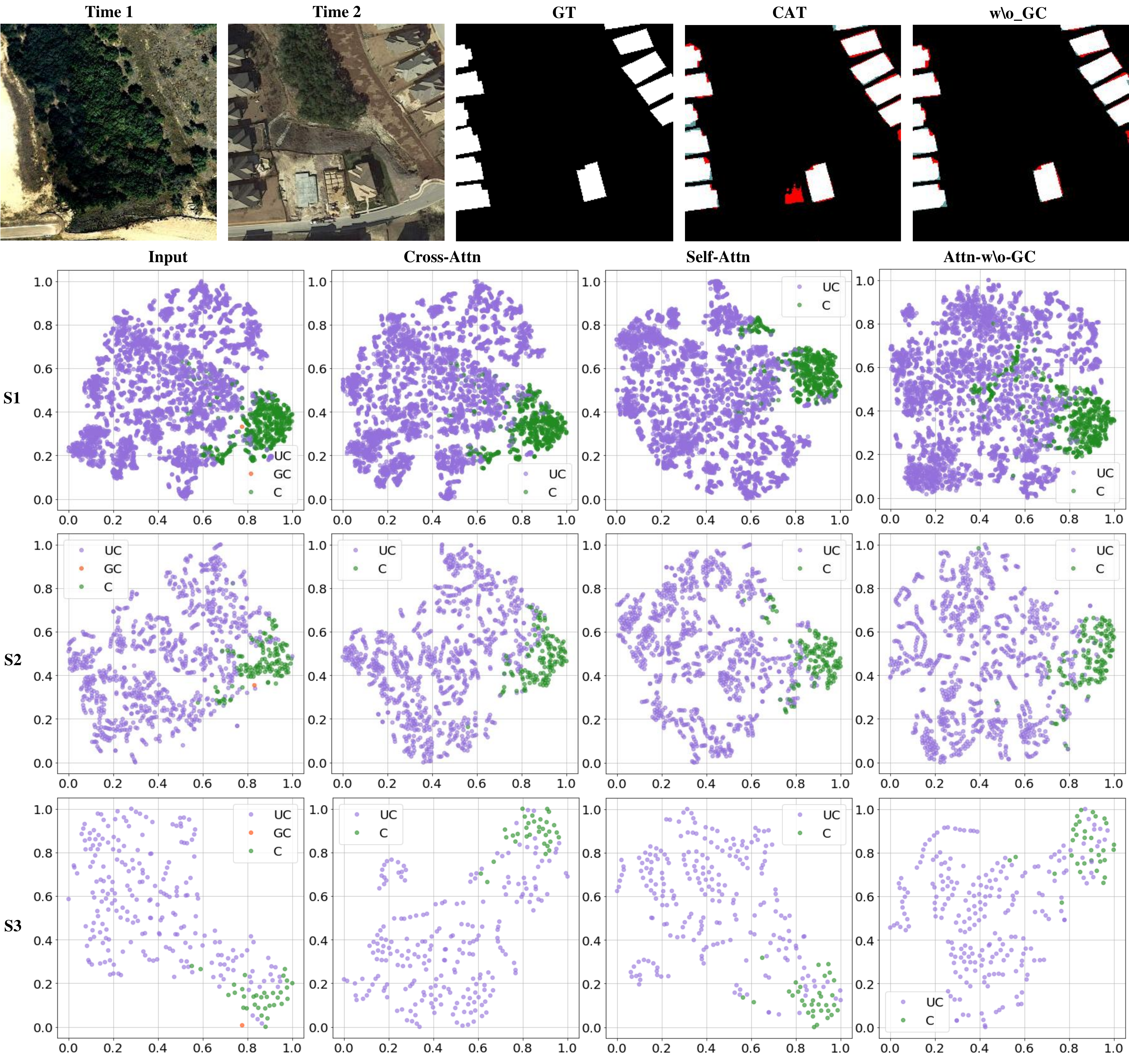}
		\label{fig_tsne_3}}
	\subfloat[]{\includegraphics[width=0.49\linewidth]{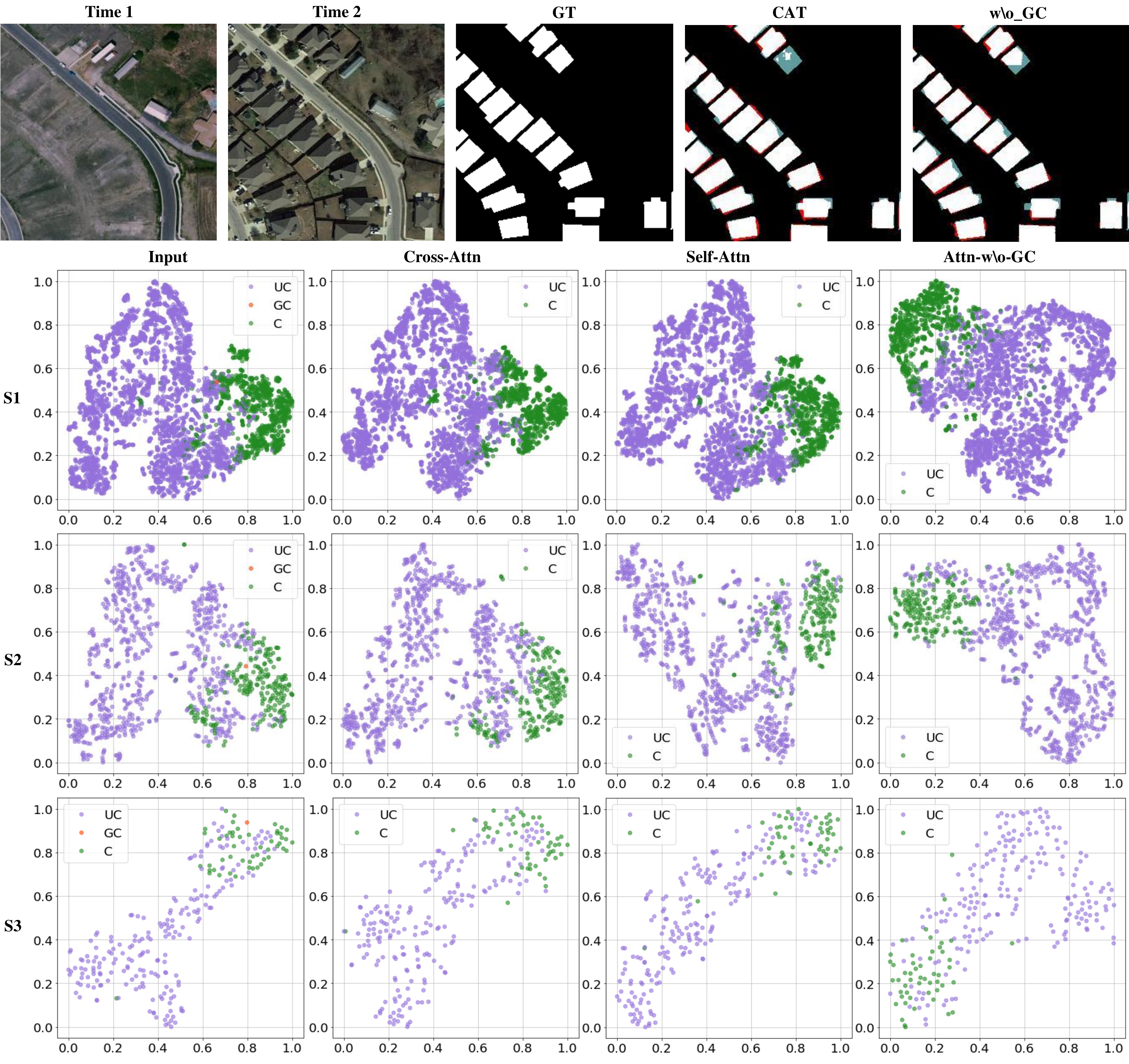}
		\label{fig_tsne_4}}
	\caption{Visualization of two CAT's failure samples.}
	\label{fig_failure}
\end{figure*}

We show two failure examples of CAT in Fig. \ref{fig_failure} following the visualization settings in Fig. \ref{fig_ablation}. In the features of the "Input" column in Fig. \ref{fig_tsne_3}, the GC representations in the first two stages are also close to some noisy pixels. Then, the cosine cross-attention and self-attention layers gather these noisy pixels into the changed pixels cluster. The clusters in the CAT features of the third stage are more isolated, while the changed pixels and noisy pixels are still together. It demonstrates a risk of CAT: if the GC representation is also close to noisy pixels, the CAT might push them into changed pixels in the difference feature space. Consequently, these noisy pixels are misclassified. In Fig. \ref{fig_tsne_4}, CAT-Siam-T misclassifies more changed pixels as unchanged. We believe this is because the GC representation is close to some unchanged pixels but far away from some changed ones, and these changed pixels are pushed away from other changed ones. It uncovers another risk of CAT: when GC representation and some changed pixels differ greatly, CAT may augment this difference, and as a result, these changed pixels are miscategorized as unchanged. Overall, if the GC representation in the difference feature domain is very similar/related to some noisy pixels, or if it only learns a part of changes knowledge, it may lead to performance degradation. So, there is still much room for improvement in GC learning methods.

Additionally, we tried to deploy CAT in multi-class change detection task \cite{varghese2018changenet}, but the performance was inferior, and there are high Recall but low Precision. We attribute it to the GC representation extracts tends to learn generalized features of all kinds of changes. When it comes to determining in detail the type of changes, this becomes a hindrance.

\subsection{Analysis of Resource Consumption}
\begin{table*}[h]
	\caption{The number of parameters (millions) and GFLOPS of our CD models. "ChnModulation" stands for the convolution layer in CAT-Siam-R that adjusts the number of channels of features.}
	\label{tab_para}
	\centering
	\footnotesize
	\begin{tabular}{@{}ccccccccc@{}}
		\toprule
		& Model      & Backbone & ChnModulation    & IDFs & CAT  & DUD  & Classifier  & Total \\ \midrule
		\multirow{2}{*}{Params.(M)} & CAT-Siam-T & 13.71    & N/A   & 3.49 & 6.24 & 1.11 & 0.22  & 24.77 \\ \cmidrule(l){2-9} 
		& CAT-Siam-R & 2.783    & 0.13  & 3.49 & 6.24 & 1.11 & 0.22  & 13.97 \\ \midrule
		\multirow{2}{*}{GFLOPS(G)}   & CAT-Siam-T & 15.474   & N/A   & 2.04 & 3.21 & 0.93 & 14.56 & 36.21 \\ \cmidrule(l){2-9} 
		& CAT-Siam-R & 3.37     & 0.153 & 2.04 & 3.21 & 0.93 & 14.56 & 24.57 \\ \bottomrule
	\end{tabular}
\end{table*}
Table \ref{tab_para} lists the number of parameters and the amount of calculation of CAT-Siam-T and CAT-Siam-R. For CAT-Siam-T, the local window attention-based Transformer backbone network has the most parameters and is the most computationally intensive. CAT has the next largest number of parameters, but its calculation amount is not so large, i.e., 3.21 GFLOPS. The amount of the classifier's computation is notable, although its number of parameters is small. This is because the classifier works on feature maps with a large size of $ 256 \times 256 $. CAT-Siam-R has a smaller amount of parameters and calculations because its ResNet18 backbone network is resource-save compared to Transformers. Note that we use a convolution layer to adjust the number of channels of features outputted by ResNet in each stage of CAT-Siam-R. Concretely, it transforms the number of channels from 64, 128, and 256 to 96, 192, and 384, respectively, and it has a few parameters (0.13M) and calculations (0.153 GFLOPS). Overall, the consumption of resources by CAT is acceptable. The backbone network integrated with CAT can be selected according to different resource conditions and performance requirements.

Table \ref{tab_compare_para} lists the number of parameters and GFLOPS of recent CD methods. Comparing CAT-Siam-T and ChangeFormer, our local window attention-based model is more efficient. BIT extracts image features from CNNs, and its Transformer module uses few tokens, so the amount of parameters and computation is low. DESSN employs Siamese UNet as the backbone network, which has fewer parameters than our models, but is more computationally intensive. In short, CAT-Siam-T and CAT-Siam-R achieve SOTA performance with a moderate number of parameters and calculations.

\begin{table}[h]
	\caption{Comparison of the number of parameters (millions) and GFLOPS of different methods.}
	\label{tab_compare_para}
	\centering
	\footnotesize
	\begin{tabular}{@{}ccc@{}}
		\toprule
		Models       & GFLOPS  & Parameters. \\ \midrule
		IFNet        & 112.15G & 43.5M  \\
		DESSN        & 36.75G  & 19.35M \\
		BIT          & 10.60G  & 12.40M \\
		ChangeFormer & 138.45G & 41.03M \\
		DR-TANet     & 6.72G   & 33.39M \\
		C-3PO        & 60.14G  & 40.54M \\
		CAT-Siam-R   & 24.57G  & 13.97M \\
		CAT-Siam-T   & 36.21G  & 24.77M \\ \bottomrule
	\end{tabular}
\end{table}

\section{Conclusion}
We propose to learn the generalized changes representation and develop a novel Changes-Aware Transformer (CAT) for CD. It directly perceives changed pixels and refines the pixels' features in the difference feature space through the cosine cross-attention mechanism, which is a more explicit and effective way to make changed pixels more similar to each other and more distinct from those unchanged ones. CAT can easily be integrated into Transformers or CNNs backbone, as well as the existing methods, and brings consistent improvements. Experiments on four change detection data sets with different scenes and objects demonstrate our approach achieves state-of-the-art results. Analyses of the advantages and disadvantages of CAT are conducted thoroughly. Valuable empirical results for CD are presented. Future work includes improving the GC representation and extending the CAT to the multi-class change detection task.


%



%
%

\ifCLASSOPTIONcaptionsoff
\newpage
\fi



\small\bibliographystyle{IEEEtranN}
\bibliography{IEEEabrv,cd_ref}
\end{document}